\documentclass [11pt] {article}
\usepackage{latexsym}
\usepackage{graphicx}

\RequirePackage{natbib}
\RequirePackage{url}

\newtheorem{theorem}{Theorem}
\newtheorem{lemma}{Lemma}

\title{Online Learning with Continuous Ranked Probability Score}
\author{{Vladimir V'yugin}
\\
{\small Institute for Information Transmission Problems}
\\
{\small Skolkovo Institute of Science and Technology}
\\
{\small (Moscow, Russia)}
\\
{\small e-mail vyugin@iitp.ru}
\\
{Vladimir Trunov}
\\
{\small Institute for Information Transmission Problems}
\\
{\small (Moscow, Russia)}
\\
{\small e-mail trunov@iitp.ru}}

\date{}

\def\argmin[{{\rm argmin}}

\def\y{{\bf y}}

\def\p{{\bf p}}
\def\q{{\bf q}}

\def\w{{\bf w}}

\def\0{{\bf 0}}
\def\c{{\bf c}}

\def\e{{\bf e}}

\def\Subst{{\rm Subst}}

\def\0{{\bf 0}}
\def\f{{\bf f}}

\usepackage{bm}

\begin{document}

\maketitle

\def\CRPS{{\rm CRPS}}
\def\QUANT{{\rm QUANT}}

\begin{abstract}%

We consider the case where several competing methods produce online predictions
in the form of probability distribution functions.
The dissimilarity between a probability forecast and an outcome is measured by
a loss function (scoring rule). Popular example of scoring rule for continuous
outcomes is Continuous Ranked Probability Score ($\CRPS$).
In this paper the problem of combining probabilistic forecasts is considered in
the prediction with expert advice framework. We show that $\CRPS$ is a mixable
loss function and then the time-independent upper bound for the regret of the
Vovk's aggregating algorithm using $\CRPS$ as a loss function can be
obtained. Also, we incorporate in this scheme a "smooth" version of the method of specialized
experts which allows us to make more flexible and accurate predictions.
We present the results of numerical experiments illustrating the proposed methods.
\end{abstract}


\section{Introduction}

Probabilistic forecasts in the form of probability distributions over future events have
become popular in several fields including meteorology, hydrology, economics,
demography (see discussion in~\citealt{JKL2018}).
Probabilistic predictions are used in the theory of conformal predictions, where
a predictive distribution that is valid under a nonparametric assumption
can be assigned to any forecasting algorithm (see~\citealt{VSMX2018}).

The dissimilarity between a probability forecast and an outcome is measured by
a loss function (scoring rule). Popular example of scoring rule for continuous
outcomes is Continuous Ranked Probability Score ($\CRPS$).
$$
\CRPS(F,y)=\int (F(u)-H(u-y))^2 du,
$$ 
where $F(u)$ is a probability distribution function, $y$ is an outcome -- a real
number, and $H(x)$ is the Heaviside function: $H(x)=0$ for $x<0$ and $H(x)=1$ 
for $x\ge 0$

We consider the case where several competing methods produce online predictions
in the form of probability distribution functions. These predictions can lead to
large or small losses. Our task is to combine these forecasts into one optimal forecast,
which will lead to the smallest possible loss in the framework of the available past
information.

We solve this problem in the prediction with expert advice (PEA) framework. We consider
the game-theoretic on-line learning model in which a learner (aggregating)
algorithm has to combine predictions from a set of $N$ experts
(see e.g.~\citealt{LiW94},~\citealt{FrS97},~\citealt{Vov90},~\citealt{VoV98},
~\citealt{cesa-bianchi} among others).

In contrast to the standard PEA approach, we consider the case where each expert
presents probability distribution functions rather than a point prediction.
The learner presents his forecast also in a form of probability distribution function
computed using probabilistic predictions presented by the experts.

In online setting, at each time step $t$ any expert issues
a probability distribution as a forecast. The aggregating algorithm combines these
forecasts into one aggregated forecast, which is a probability distribution function.
The effectiveness of the aggregating algorithm
on any time interval $[1,T]$ is measured by the regret which is the difference between
the cumulated loss of the aggregating algorithm and the cumulated loss of the best expert
suffered on first $T$ steps.

There are a lot of papers on probabilistic predictions and on $\CRPS$ scoring rule
(some of them are \citealt{Bri1950}, \citealt{Bro2007}, \citealt{Bro2008}, \citealt{Bro2012},
\citealt{Eps69}, \citealt{JKL2018}, \citealt{RGBP2005}).
Most of them referred to the ensemble interpretation models.
In some cases, experts use for their predictions probability distributions functions
(data models) which are defined explicitly in an analytic form. In this paper we propose
the rules for aggregation of such the probability distributions functions.
We present the exact formulas for direct calculation of the aggregated probability
distribution function given probability distribution functions presented by the experts.



In this paper we obtain a tight upper bound of the regret for a special
case when the outcomes and the probability distributions are located in a finite
interval $[a,b]$ of real line. We show that the loss function $\CRPS(F,y)$ is mixable in
sense of~\citet{VoV98} and apply the Vovk's aggregating algorithm to
obtain the time-independent upper bound $\frac{b-a}{2}\ln N$ for the regret.\footnote{
The complete definitions are given in Section~\ref{app-1}.}


The application we will consider below in Section~\ref{exp-1}
(which is the sequential short-term (one-hour-ahead) forecasting of
probability distribution function of electricity consumption) will take place in a variant
of the basic problem of prediction with expert advice called prediction with specialized
(or sleeping) experts. At each round only some of the experts output a prediction while
the other ones are inactive. Each expert is expected to provide accurate forecasts
mostly in given external conditions, that can be known beforehand. For instance,
in the case of the prediction of electricity consumption, experts can be specialized
to a season, temperature, to time of day.

In Section~\ref{main-1} we prove that the $\CRPS$ function is mixable and then
all machinery of the~\citet{VoV98} aggregating algorithm (AA) and of the exponentially
weighted average forecaster (WA) can be applied (see~\citealt{cesa-bianchi}).

In Section~\ref{main-1} we present a method for computing online
the aggregated probability distribution function given the probability distribution
functions presented by the experts and prove a time-independent bound for the regret of
the proposed algorithm.

The method of specialized experts was first proposed by~\citet{Freu-2} and
further developed by~\citet{ChV2009},~\citet{electricity},~\citet{KAS2015}.
With this approach, at each step $t$, a set of specialized experts
$E_t\subseteq\{1,\dots, N\}$ be given. A specialized expert $i$ issues its forecasts
not at all steps $t=1,2,\dots$, but only when $i\in E_t$. At any step, the aggregating
algorithm uses forecasts of only ``active (non-sleeping)'' experts.

The second contribution of this paper is that we have incorporated into
the aggregating algorithm a smooth generalization of the method of specialized experts
(Sections~\ref{v-s-1} and~\ref{CRPS-3}).
At each time moment $t$, we complement the expert $i$ forecast by a confidence level
which is a real number $p_{i,t}\in [0,1]$.
In particular, $p_{i,t}=1$ means that the forecast of the expert $i$ is used in full,
whereas in the case of $p_{i,t}=0$ it is not taken into account at all (the expert sleeps).
In cases where $0<p_{i,t}<1$ the expert's forecast is partially taken into account.
For example, when moving from one season to another, an expert tuned to the previous
season gradually loses his predictive abilities.
The dependence of $p_{i,t} $ on values of exogenous parameters can be predetermined
by a specialist in the domain or can be constructed using regression analysis on
historical data.

We demonstrate the effectiveness of the proposed methods in Section~\ref{exp-1}, where
the results of numerical experiments with synthetic and real data are presented.

\section{Preliminaries}\label{app-1}

In this section we present the main definitions and the auxiliary results
of the theory of prediction with expert advice, namely, learning with mixable
loss functions.

\subsection{Online learning}\label{online-learning-1}

Let a set of outcomes $\Omega$ and a set $\Gamma$ of forecasts (decision space) be
given.\footnote{In general, these sets can be of arbitrary nature.
We will specify them when necessary.}
We consider the learning with a loss function $\lambda(f,y)$, where
$f\in\Gamma$ and $y\in\Omega$.
Let also, a set $E$ of experts be given. For simplicity we assume that $E=\{1,\dots ,N\}$.

In PEA approach the learning process is represented as a game. The experts and the learner
observe past real outcomes generated online by some adversarial mechanism (called nature)
and present their forecasts. After that, a current outcome is revealed by the nature.


At any round $t=1,2,\dots$ each expert $i\in E$ presents a forecast $f_{i,t}\in\Gamma$, then
the learner presents its forecast $f_t\in\Gamma$, after that, an outcome $y_t\in\Omega$
will be revealed. Each expert $i$ suffers the loss $\lambda(f_{i,t},y_t)$ and the learner
suffers the loss $\lambda(f_t,y_t)$; see Protocol 1 below.

\smallskip

{\bf Protocol 1}

{\small
\medskip\hrule\hrule\medskip

\medskip
\noindent{\bf FOR} $t=1,\dots ,T$
\begin{enumerate}
\item Receive the experts' predictions $f_{i,t}$, where $1\le i\le N$.
\item Present the learner's forecast $f_t$.
\item Observe the true outcome $y_t$ and compute the losses
$\lambda(f_{i,t},y_t)$ of the experts and the loss $\lambda(f_t,y_t)$ of the learner.
\end{enumerate}

\noindent \hspace{2mm}{\bf ENDFOR}
\medskip\hrule\hrule\medskip
}
\smallskip

Let $H_T=\sum\limits_{t=1}^T \lambda(f_t,y_t)$ be the cumulated loss of the learner and
$L^i_T=\sum\limits_{t=1}^T \lambda(f_{i,t},y_t)$ be the cumulated loss of an expert $i$.
The difference $R^i_T=H_T-L^i_T$ is called
regret with respect to an expert $i$ and $R_T=H_T-\min_i L^i_T$ is the regret
with respect to the best expert. The goal of the learner is to minimize the regret.

\subsection{Aggregating Algorithm (AA)}\label{AA-1}

The Vovk's Aggregating algorithm (\citealt{Vov90},~\citealt{VoV98}) is the base
algorithm for computing the learner predictions. This algorithm starting from
the initial weights $w_{i,1}$ (usually $w_{i,1}=\frac{1}{N}$ for all $i$)
assign weights $w_{i,t}$ for the experts $i\in E$ using the weights
update rule:
\begin{eqnarray}
w_{i,t+1}=w_{i,t}e^{-\eta\lambda(f_{i,t},y_t)}\mbox{ for } t=1,2,\dots,
\label{wei-up-1}
\end{eqnarray}
where $\eta>0$ is a learning rate. The normalized weights are defined
\begin{eqnarray}
w^*_{i,t}=\frac{w_{i,t}}{\sum\limits_{j=1}^N w_{j,t}}.
\label{weight-update-1}
\end{eqnarray}
The main tool of AA is a superprediction function
\begin{eqnarray}
g_t(y)=-\frac{1}{\eta}\ln\sum\limits_{i=1}^N e^{-\eta\lambda(f_{i,t},y)}w^*_{i,t}.
\label{superpred-1}
\end{eqnarray}
We consider probability distributions $\q=(q_1,\dots ,q_N)$ on
the set $E$ of the experts: $\sum\limits_{i=1}^N q_i=1$ and $q_i\ge 0$ for all $i$.
By~\citet{VoV98} a loss function is called $\eta$-mixable if for any probability
distribution $\q$ on the set $E$ of experts and for any predictions
$\c=(c_1,\dots ,c_N)$ of the experts there exists a forecast $f$ such that
\begin{eqnarray}
\lambda(f,y)\le g(y)\mbox{ for all } y,
\label{mix-1}
\end{eqnarray}
where
\begin{eqnarray}
g(y)=-\frac{1}{\eta}\ln\sum\limits_{i=1}^N e^{-\eta\lambda(c_i,y)}q_i.
\label{superpred-1a}
\end{eqnarray}
We fix some rule for calculating a forecast $f$ and write
\begin{eqnarray}
f=\Subst(\c,\q).
\label{AA_rule-1}
\end{eqnarray}
The function $\Subst$ is called the substitution function.

As follows from (\ref{mix-1}) and (\ref{superpred-1a}) if a loss function
$\lambda(f,y)$ is $\eta$-mixable then the loss function
$c\lambda(f,y)$ is $\frac{\eta}{c}$-mixable for any $c>0$.

{\bf Regret analysis for AA.}
Assume that a loss function $\lambda(f,y)$ is $\eta$-mixable. Let
$\w^*_t=(w^*_{1,t},\dots ,w^*_{N,t})$ be the normalized weights and
$\f_t=(f_{1,t},\dots ,f_{N,t})$ be the experts' forecasts
at step $t$. Define in Protocol 1 the learner's forecast $f_t=\Subst(\f_t,\w^*_t)$.
By (\ref{mix-1}) $\lambda(f_t,y_t)\le g_t(y_t)$ for all $t$, where
$g_t(y)$ is defined by (\ref{superpred-1}).

Let $H_T=\sum\limits_{t=1}^T \lambda(f_t,y_t)$ be the cumulated loss of the
learner and $L^i_T=\sum\limits_{t=1}^T \lambda(f_{i,t},y_t)$ be the
cumulated loss of an expert $i$. By definition
$g_t(y_t)=-\frac{1}{\eta}\ln\frac{W_{t+1}}{W_t}$, where 
$W_t=\sum\limits_{i=1}^N w_{i,t}$ and $W_1=1$.
By the weight update rule (\ref{wei-up-1}) we obtain
$w_{i,t+1}=\frac{1}{N}e^{-\eta L^i_t}$.

By telescoping, we obtain the time-independent bound
\begin{eqnarray}
H_T\le\sum\limits_{t=1}^T g_t(y_t)=-\frac{1}{\eta}\ln W_{T+1}\le
L^i_T+\frac{\ln N}{\eta}
\label{prop-1}
\end{eqnarray}
for any expert $i$. Therefore, there is a strategy for the learner that guarantees
$R_T\le\frac{\ln N}{\eta}$ for all $T$.

{\bf Exponential concave loss functions.}
Assume that the set of all forecasts form a linear space. In this case,
the mixability is a generalization of the notion of the exponential concavity.
A loss function $\lambda(f,y)$ is called $\eta$-exponential concave if
for each $\omega$ the function $\exp (-\eta\lambda(f,y))$ is concave by
$f$ for any $y$ (see~\citealt{cesa-bianchi}). For exponential concave loss function
the inequality $\lambda(f,y)\le g(y)$ holds for all $y$ by definition, if the
forecast of the learner is computed using the weighted average (WA)
of the experts predictions:
\begin{eqnarray}
f=\sum\limits_{i=1}^N q_i f_i,
\label{mix-forecast-2}
\end{eqnarray}
where $\q=(q_1,\dots ,q_N)$ is a probability distribution on the set of experts, and
$f_1,\dots ,f_N$ are theirs forecasts.

For exponential concave loss function and the game defined by Protocol 1,
where the learner's forecast is computed by (\ref{mix-forecast-2}), we also have
the time-independent bound (\ref{prop-1}) for the regret.

{\bf Square loss function.} The important special case is $\Omega=\{0,1\}$
and $\Gamma=[0,1]$. The square loss function
$\lambda(\gamma,\omega)=(\gamma-\omega)^2$ is $\eta$-mixable loss function
for any $0<\eta\le 2$, where $\gamma\in [0,1]$ and $\omega\in\{0,1\}$.\footnote{
In what follows $\omega_t$ denotes a binary outcome.}
In this case, at any step $t$, the corresponding forecast $f_t$ (in Protocol 1)
can be defined as
\begin{eqnarray}
f_t=\Subst(\f_t,\w^*_t)=\frac{1}{2}-
\frac{1}{2\eta}\ln\frac{\sum\limits_{i=1}^N w^*_{i,t} e^{-\eta\lambda(f_{i,t},0)}}
{\sum\limits_{i=1}^N w^*_{i,t}e^{-\eta\lambda(f_{i,t},1)}},
\label{mix-forecast-1}
\end{eqnarray}
where $\f_t=(f_{1,t},\dots ,f_{N,t})$ is the vector of the experts' forecasts and
$\w^*_t=(w^*_{1,t},\dots ,w^*_{N,t})$ is the vector of theirs normalized weights
defined by (\ref{wei-up-1}) and~(\ref {weight-update-1}).
We refer the reader for details to \citet{Vov90},~\citet{VoV98}, and~\citet{Vov2001}.

The square loss function $\lambda(f,\omega)=(f-\omega)^2$, where $\omega\in\{0,1\}$
and $f\in [0,1]$, is $\eta$-exponential concave for any $0<\eta\le\frac{1}{2}$
(see~\citealt{cesa-bianchi}).

\section{AA for experts with confidence}\label{v-s-1}

In this section we consider an extended setting. At each time moment $t$
the forecasts $\f_t= (f_{1,t},\dots f_{N,t})$ of the experts
and confidence levels $\p_t=(p_{1,t},\dots ,p_{N,t})$ of these forecasts are
revealed.

Each confidence level is a number between 0 and 1. If  $p_{i,t}=0$ then
the corresponding expert ``sleeps'' at step $t$ and its forecast is not taken
into account. If $p_{i,t}<1$ then we will use the forecast $f_{i,t}$ only partially.
Confidence levels can be set by the expert itself or by the learner.
\footnote{The setting of prediction with experts that report their confidences as a number
in the interval $[0,1]$ was first studied by~\citet{BlM2007} and further developed
by~\citet{CBMS2007}.}

To take into account confidence levels, we use the fixed point method by~\citet{ChV2009}.
We consider any confidence level $p_{i,t}$ as a probability distribution
$\p_{i,t}=(p_{i,t}, 1-p_{i,t})$ on a two element set.
Define the auxiliary forecast:
\[
\tilde f_{i,t}=
\left\{
    \begin{array}{l}
      f_{i,t}\mbox{ with probability } p_{i,t},
    \\
      f_t\mbox{ with probability } 1-p_{i,t},
    \end{array}
  \right.
\]
where $f_t$ is a forecast of the learner. 

First, we provide a justification of the algorithm presented below.
Our goal is to define the forecast $f_t$ such that
\begin{eqnarray}
e^{-\eta\lambda(f_t,y)}\ge\sum_{i=1}^N
E_{\p_{i,t}}[e^{-\eta\lambda(\tilde f_{i,t},y)}]w_{i,t}
\label{cond-1a}
\end{eqnarray}
for each $y$. Here $E_{\p_{i,t}}$ is the mathematical expectation
with respect to the probability distribution $\p_{i,t}$. Also,
$w_{i,t}$ is the weight of the expert $i$ accumulated at the end of step $t$.

We rewrite inequality (\ref{cond-1a}) in a more detailed form:
\begin{eqnarray}
e^{-\eta\lambda(f_t,y)}\ge
\sum_{i=1}^N E_{\p_{i,t}}[e^{-\eta\lambda(\tilde f_{i,t},y)}]w_{i,t}=
\label{cond-1}
\\
\sum_{i=1}^N p_{i,t}w_{i,t}e^{-\eta\lambda(f_{i,t},y)}+
e^{-\eta\lambda(f_t,y)}
\left(1-\sum_{i=1}^N p_{i,t}w_{i,t}\right)
\label{cond-2}
\end{eqnarray}
for all $\omega$.
Therefore, the inequality (\ref{cond-1a}) is equivalent to the inequality
\begin{eqnarray}
e^{-\eta\lambda(f_i,y)}\ge\sum_{i=1}^N
w^*_{i,t}e^{-\eta\lambda(f_{i,t},y)},
\label{for-1ba}
\end{eqnarray}
where
\begin{eqnarray}\label{for-1bb}
w^*_{i,t}=\frac{p_{i,t} w_{i,t}}{\sum_{j=1}^N p_{j,t} w_{j,t}}.
\end{eqnarray}
According to the rule (\ref{AA_rule-1}) for computing the forecast of AA
define $f_t=\Subst(\f_t,\w^*_t)$. Then
(\ref{for-1ba}) and its equivalent (\ref{cond-1}) are valid.
Here $\Subst$ is the substitution function, $\w^*_t=(w^*_{i,1},\dots , w^*_{i,N})$ and
$\f_t=(f_{1,t},\dots f_{i,N})$.

Now, we present the analogue of Protocol 1 for AA with confidence.

\smallskip

{\bf Protocol 1a}

{\small
\medskip\hrule\hrule\medskip

\medskip
\noindent{\bf FOR} $t=1,\dots ,T$
\begin{enumerate}
\item Receive the experts' predictions $f_{i,t}$ and confidence levels $p_{i,t}$,
where $1\le i\le N$.
\item Present the learner's forecast $f_t=\Subst(\f_t,\w^*_t))$, where
normalized weights $\w^*_t=(w^*_{1,t},\dots ,w^*_{N,t})$ are defined by (\ref{for-1bb}).
\item Observe the true outcome $y_t$ and compute the losses
$l_{i,t}=\lambda(f_{i,t},y_t)$ of the experts and the loss $\lambda(f_t,y_t)$ of the learner.
\item Update the weights (of the virtual experts) by the rule 
\begin{eqnarray}
w_{i,t+1}=w_{i,t}e^{-\eta (p_{i,t}\lambda(f_{i,t},y_t)+(1-p_{i,t})\lambda(f_t,y_t))}
\label{for-1b-2}
\end{eqnarray}
\end{enumerate}

\noindent \hspace{2mm}{\bf ENDFOR}
\medskip\hrule\hrule\medskip
}
\smallskip

Let $h_t=\lambda(f_t,y_t)$ be the loss of the learner at time moment $t$,
$\tilde l_{i,t}=\lambda(\tilde f_{i,t},y_t)$ be the estimated loss of an expert $i$,
and $\hat l_{i,t}=E_{\p_{i,t}}[\tilde l_{i,t}]$ be its expectation.
By virtual expert we mean the expert which suffers the loss $\hat l_{i,t}$.

Since by definition $\hat l_{i,t}=p_{i,t}l_{i,t}+(1-p_{i,t})h_t$,
we have $h_t-\hat l_{i,t}=p_{i,t}(h_t-l_{i,t})$. We call the last quantity
discounted excess loss with respect to an expert $i$ at a time moment $t$ and
we will measure the performance of our algorithm by the cumulative discounted excess
loss with respect to any expert $i$.
\begin{theorem}\label{main-3fa}
For any ${1\le i\le N}$, the following upper bound
for the cumulative excess loss (discounted regret) holds true:
\begin{eqnarray}
\sum\limits_{t=1}^T p_{i,t}(h_t-l_{i,t})\le\frac{\ln N}{\eta}.
\label{mTT-1afaa}
\end{eqnarray}
\end{theorem}
{\it Proof}.
By convexity of the exponent the inequality (\ref{cond-1a}) implies
\begin{eqnarray}
e^{-\eta\lambda(f_t,y)}\ge\sum_{i=1}^N e^{-\eta E_{\p_{i,t}}
[\lambda(\tilde f_{i,t},y)]}w^*_{i,t}=
\sum_{i=1}^N e^{-\eta\hat l_{i,t}}w^*_{i,t}.
\label{for-1b-2s}
\end{eqnarray}
Let $m_t=-\frac{1}{\eta}\ln\sum_{i=1}^N w^*_{i,t}e^{-\eta\hat l_{i,t}}$.
By (\ref{for-1b-2s}) $h_t\le m_t$. Rewrite the update rule (\ref{for-1b-2}) as
\begin{eqnarray}
w_{i,t+1}=w_{i,t}e^{-\eta\hat l_{i,t}}.
\label{for-1b-2a}
\end{eqnarray}
Recall that $W_T=\sum_{t=1}w_{i,t}$, $W_1=1$ and $m_t=\frac{1}{\eta}\ln\frac{W_{t+1}}{W_t}$.
As in (\ref{prop-1}), using (\ref{for-1b-2s}) and (\ref{for-1b-2a}), we obtain
\begin{eqnarray*}
\sum_{t=1}^T h_t\le\sum_{t=1}^T m_t=-\frac{1}{\eta}\ln W_{T+1}\le
\sum_{t=1}^T \hat l_{i,t}+\frac{\ln N}{\eta}
\end{eqnarray*}
for any $i$.
Since $h_t-\hat l_{i,t}=p_{i,t}(h_t-l_{i,t})$, the inequality (\ref{mTT-1afaa})
follows.~QED


\section{Aggregation of probability forecasts}\label{main-1}

\subsection{$\CRPS$ loss function}\label{CRPS-1}

Let in Protocol 1 the set of outcomes be an interval $\Omega=[a,b]$ of the real line
for some $a<b$ and the set of forecasts $\Gamma$ be a set of all
probability distribution functions $F:[a,b]\to [0,1]$.\footnote{
A probability distribution function is a non-decreasing function
$F(y)$ defined on this interval such that $F(a)=0$ and $F(b)=1$.
Also, it is left-continuous and has the right limit at each point.}

The quality of the prediction $F$ in view of the actual outcome $y$
is often measured by the continuous ranked probability score (loss function)
\begin{equation}
\CRPS(F,y)=\int_a^b (F(u)-H(u-y))^2 du,
\label{crps-1}
\end{equation}
where $H(x)$ is the Heaviside function: $H(x)=0$ for $x<0$ and $H(x)=1$ for $x\ge 0$ 
(\citet{Eps69},~\citealt{MaW76}, etc).

For simplicity in this definition, we consider integration over a finite 
interval.
Such definition is closer to practical applications and 
allows a more elementary theoretical analysis.
More general definition includes a density $\mu(u)$ and integration over the real line: 
\begin{equation}
\CRPS(F,y)=\int_{-\infty}^{+\infty} (F(u)-H(u-y))^2 \mu(u)du.
\label{crps-1a}
\end{equation}
The definition (\ref{crps-1}) is a special case of this definition (up to a factor), 
where $\mu(u)=\frac{1}{b-a}$ for $u\in [a,b]$ and $\mu(u)=0$ otherwise.
In can be proved that the function (\ref{crps-1a}) is $\eta$-mixable for $0<\eta\le 2$
and $\eta$-exponentially concave for $0<\eta\le\frac{1}{2}$ (see~\citealt{KVG2019}).

The $\CRPS$ score measures the difference between the forecast $F$ and a perfect forecast
$H(u-y)$ which puts all mass on the verification $y$.
The lowest possible value $0$ is attained
when $F$ is concentrated at $y$, and in all other cases $\CRPS(F,y)$ will be positive.

We consider a game of prediction with expert advice, where the forecasts of the experts
and of the learner are (cumulative) probability distribution functions.
At any step $t$ of the game
each expert $i\in \{1,\dots ,N\}$ presents its forecast -- a probability distribution
function $F_{i,t}(u)$ and the learner presents its forecast $F_t(u)$.\footnote{
For simplicity of presentation we consider the case where the set of the experts is finite.
In case of infinite $E$, the sums by $i$ should be replaced by integrals with respect
to the corresponding probability distributions on the set of experts. In this case
the choice of initial weights on the set of the experts is a non-trivial problem.}
After an outcome $y_t\in [a,b]$ be revealed and the experts and the learner suffer losses
$\CRPS(F_{i,t},y_t)$ and $\CRPS(F_t,y_t)$. The corresponding game of
probabilistic prediction is defined by the following protocol:

\smallskip

{\bf Protocol 2}

{\small
\medskip\hrule\hrule\medskip

\medskip
\noindent{\bf FOR} $t=1,\dots ,T$
\begin{enumerate}
\item Receive the experts' predictions -- the probability distribution
functions $F_{i,t}(u)$ for $1\le i\le N$.
\item Present the learner's forecast -- the probability distribution
function $F_t(u)$.
\item Observe the true outcome $y_t$ and compute the scores

$\CRPS(F_{i,t},y_t)=\int_a^b (F_{i,t}(u)-H(u-y_t))^2 du$
of the experts $1\le i\le N$

and the score

$\CRPS(F_t,y_t)=\int_a^b (F_t(u)-H(u-y_t))^2 du$
of the learner.
\end{enumerate}

\noindent \hspace{2mm}{\bf ENDFOR}
\medskip\hrule\hrule\medskip
}
\smallskip

The goal of the learner is to predict such that independently of which outcomes
are revealed and the experts' predictions are presented its cumulated loss
$L_T=\sum\limits_{t=1}^T\CRPS(F_t,y_t)$ is asymptotically less
than the loss $L^i_T=\sum\limits_{t=1}^T\CRPS(F_{i,t},y_t)$
of the best expert $i$ up to some regret and $L_T-\min_i L^i_T=o(T)$ as $T\to\infty$.

First, we show that $\CRPS$ loss function (and the corresponding game) is mixable.
\begin{theorem}\label{theorem-1}
The continuous ranked probability score $\CRPS(F,y)$ is $\frac{2}{b-a}$-mixable
loss function.
The corresponding learner's forecast $F(u)$ given the forecasts $F_i(u)$ of
the experts $1\le i\le N$ and a probability distribution $\q=(q_1,\dots ,q_N)$ on
the set of all experts can be computed by the rule \footnote{
Easy to verify that $F(u)$ is a probability distribution function.}
\begin{eqnarray}
F(u)=\frac{1}{2}-\frac{1}{4}\ln\frac{\sum_{i=1}^N q_i e^{-2(F_i(u))^2}}
{\sum_{i=1}^N q_i e^{-2(1-F_i(u))^2}},
\label{forecast-2}
\end{eqnarray}
\end{theorem}
{\it Proof}. 
We approximate any probability distribution function $F(u)$ by 
a piecewise-constant function that take a finite number 
of values on a uniform grid of arguments. Accordingly, the forecasts of the experts and 
of the learner will take the form of $d$-dimensional vectors. We apply AA to
the $d$-dimensional forecasts, then we consider the limit $d\to\infty$.

~\citet{Kaln2017} generalize the AA for the case of $d$-dimensional forecasts, where
$d$ is a positive integer number.
Let an $\eta$-mixable loss function $\lambda(f,y)$ be given, where $\eta>0$, $f\in\Gamma$
and $y\in\Omega$. Let $\f=(f^1,\dots ,f^d)\in\Gamma^d$ be a $d$-dimensional forecast and
$\y=(y^1,\dots ,y^d)\in\Omega^d$ be a $d$-dimensional outcome. The generalized loss
function is defined $\lambda(\f,\y)=\sum\limits_{s=1}^d\lambda(f^s,y^s)$; we call
$\lambda(f,y)$ its source function.

The corresponding (generalized) game can be presented by Protocol 1
where at each step $t$ the experts
and the learner present $d$-dimensional forecasts: at any round $t=1,2,\dots$ each
expert $i\in\{1,\dots ,N\}$ presents a vector of forecasts
$\f_{i,t}=(f^1_{i,t},\dots, f^d_{i,t})$ and the learner
presents a vector of forecasts $\f_t=(f^1_t,\dots ,f^d_t)$.
After that, a vector $\y_t=(y^1_t,\dots ,y^d_t)$ of outcomes
will be revealed and the experts and the learner suffer losses
$\lambda(\f_{i,t},\y_t)=\sum\limits_{s=1}^d\lambda(f^s_{i,t},\y^s_t)$
and
$\lambda(\f_t,\y_t)=\sum\limits_{s=1}^d\lambda(f^s_t,y^s_t)$.

~\citet{Kaln2017} proved that the generalized loss function (game) is mixable.
We rewrite this result for completeness of presentation.
\begin{lemma}
The generalized loss function
$\lambda(\f,\y)$ is $\frac{\eta}{d}$-mixable if the source loss function
$\lambda(f,y)$ is $\eta$-mixable.
\end{lemma}
{\it Proof.} Let the forecasts $\c_i=(c^1_i,\dots, c^d_i)$ of the experts $1\le i\le N$
and a probability distribution $\p=(p_1,\dots ,p_N)$ on the set of the experts be given.

Since the loss function $\lambda(f,y)$ is $\eta$-mixable, we can apply the
aggregation rule to each $s$th column $\e^s=(c^s_1,\dots ,c^s_N)$ of coordinates separately:
define $f^s=\Subst(\e^s,\p)$ for $1\le s\le d$.
Rewrite the inequality (\ref{mix-1}):
\begin{eqnarray}
e^{-\eta\lambda(f^s,y}\ge\sum\limits_{i=1}^N
e^{-\eta\lambda(c^s_i,y)}p_i \mbox{ for all } y
\label{superpred-1n}
\end{eqnarray}
for $1\le s\le d$.

Let $\y=(y^1,\dots ,y^d)$ be a vector of outcomes.
Multiplying the inequalities (\ref{superpred-1n}) for $s=1,\dots ,d$
and $y=y^s$, we obtain
\begin{eqnarray}
e^{-\eta\sum_{s=1}^d\lambda(f^s,y^s)}
\ge\prod_{s=1}^d\sum_{i=1}^N e^{-\eta\lambda(c^s_i,y^s)}p_i
\label{aggr-rule-1fuapp}
\end{eqnarray}
for all $\y=(y^1,\dots ,y^d)$.

The generalized H\"older inequality says that
$$
\|G_1G_2\cdots G_d\|_r\le\|G_1\|_{q_1}\|G_2\|_{q_2}\cdots\|G_d\|_{q_d},
$$
where $\frac{1}{q_1}+\dots +\frac{1}{q_d}=\frac{1}{r}$,  $q_s\in (0,+\infty)$
and $G_s\in L^{q_s}$ for $1\le s\le d$. Let $q_s=1$ for all $1\le s\le d$, then $r=1/d$.
Let $G_{i,s}=e^{-\eta\lambda(c^s_i,y^s)}$ for $s=1,\dots ,d$ and
$\|G_s\|_1=E_{i\sim \p}[G_{i,s}]=\sum\limits_{i=1}^N G_{i,s}p_i$.
Then
\begin{eqnarray*}
e^{-\eta\frac{1}{d}\sum_{s=1}^d\lambda(f^s,y^s)}
\ge\sum_{i=1}^N e^{-\eta\frac{1}{d}\sum\limits_{s=1}^d\lambda(c^s_i,y^s)}p_i.
\end{eqnarray*}
or, equivalently,
\begin{eqnarray}
e^{-\frac{\eta}{d}\lambda(\f,\y)}\ge\sum_{i=1}^N
e^{-\frac{\eta}{d}\lambda(\c_i,\y)}p_i
\label{ii-1}
\end{eqnarray}
for all $\y=(y^1,\dots ,y^d)$, where $\f=(f^1,\dots ,f^d)$.

The inequality (\ref{ii-1}) means that the generalized loss function
$\lambda(\f,\y)$ is $\frac{\eta}{d}$-mixable.~QED

By (\ref{wei-up-1}) the weights update rule for generalized loss function
in Protocol 1 is
\begin{eqnarray*}
w_{i,t+1}=w_{i,t}e^{-\frac{\eta}{d}\lambda(\f_{i,t},\y_t)}\mbox{ for } t=1,2,\dots,
\label{wei-up-1m}
\end{eqnarray*}
where $\eta>0$ is a learning rate for the source function. The normalized weights
$\w^*_t=(w^*_{i,t},\dots, w^*_{i,t})$
are defined by (\ref{weight-update-1}). The learner forecast $\f_t=(f^1_t,\dots ,f^d_t)$
an any round $t$ is defined: $f^s_t=\Subst(\e^s_t,\w^*_t)$ for each $s=1,\dots ,d$,
where $\e^s_t=(f^s_{1,t},\dots ,f^s_{N,t})$.


We now turn to the proof of Theorem~\ref{theorem-1}.
We approximate any probability distribution
function $F(y)$ by the piecewise-constant functions $L_d(y)$, where $d=1,2,\dots$.
Any such function $L_d$ is defined by the points $z_0,z_1,z_2,\dots ,z_d$ and
the values $f_0=F(z_0)$, $f_1=F(z_1)$, $f_2=F(z_2)$, $\dots$, $f_d=F(z_d)$,
where $a=z_0<z_1<z_2<\dots <z_d=b$ and
$0=f_0\le f_1\le f_2\le \dots \le f_d=1$. By definition $L_d(y)=f_1$ for $z_0\le y<z_1$,
$L_d(y)=f_2$ for $z_1\le y<z_2$, $\dots$, $L_d(y)=f_d$ for $z_{d-1}\le y<z_d$.
Also, assume that $z_{i+1}-z_i=\Delta$ for all $0\le i<d$.
By definition $\Delta=\frac{b-a}{d}$.
We have
\begin{eqnarray}
\left|\CRPS(F,y)-\CRPS(L_d,y)\right|\le
\nonumber
\\
\int_a^y (L_d^2(u)-F^2(u))du+
\int_y^b ((1-F(u))^2-(1-L_d(u))^2)du\le 2\Delta
\label{appr-2}
\end{eqnarray}
for any $y$, since each integral is bounded by $\Delta$.
Also, we take into account that by definition $F(u)\le L_d(u)$ for all $u$.

Define an auxiliary representation of $y$, which is a binary variable
$\omega^s_y=1_{z_s\ge y}\in\{0,1\}$ for $1\le s\le d$ and
${\bm\omega}_y=(\omega^1_y,\dots,\omega^d_y)$, where $1_{z_s\ge y}=H(z_s-y)$.

Consider any $y\in [a,b]$. Easy to see that for each $1\le s\le d$
the uniform measure of all $u\in [z_{s-1},z_s]$
such that $1_{z_s\ge y}\not =1_{u\ge y}$ is less or equal to $\Delta$ if
$y\in [z_{s-1},z_s]$ and $1_{z_s\ge y}=1_{u\ge y}$ for all $u\in [z_{s-1},z_s]$
otherwise. Since $0\le f_s\le 1$ for all $s$, this implies that
\begin{eqnarray}
\left|\CRPS(L_d,y)-\Delta\sum_{s=1}^d (f_s-\omega^s_y)^2\right|\le 2\Delta
\label{appr-1}
\end{eqnarray}
for all $y$. Let us study the generalized loss function
\begin{eqnarray}
\lambda(\f,{\bm\omega})=\Delta\sum_{s=1}^d (f_s-\omega^s)^2,
\label{vf-1}
\end{eqnarray}
where $\f=(f_1,\dots ,f_d)$, ${\bm\omega}=(\omega^1,\dots,\omega^d)$ and
$\omega^s\in\{0,1\}$ for $1\le s\le d$.

The key observation is that the deterioration of the learning rate for the
generalized loss function (it gets divided by the dimension $d$ of vector-valued forecasts)
is exactly offset by the decrease in the weight of each component of the vector-valued
prediction as the grid-size decreases.

Since the square loss function $\lambda(f,\omega)=(\gamma-\omega)^2$ is
$2$-mixable, where $f\in [0,1]$ and $\omega\in\{0,1\}$, by results of
Section~\ref{app-1} the corresponding generalized loss function
$\sum_{s=1}^d (f_s-\omega^s)^2$ is $\frac{2}{d}$-mixable and then the loss function
(\ref{vf-1}) is $\frac{2}{d\Delta}=\frac{2}{b-a}$-mixable independently of that
grid-size is used.\footnote{This also means that in numerical experiments,
when calculating forecasts of the learner, we can use the same learning rate,
regardless of the accuracy of the presentation of expert forecasts.}

Let $F_i(u)$ be the probability distribution functions presented by the experts
$1\le i\le N$ and $\f_i=(f^1_i,\dots ,f^d_i)$, where $f^s_i=F^i(z_s)$ for $1\le s\le d$.
By (\ref{ii-1})
\begin{eqnarray}
e^{-\frac{2}{(b-a)}\lambda(\f,{\bm\omega})}\ge\sum_{i=1}^N
e^{-\frac{2}{b-a}\lambda(\f_i,{\bm\omega})}q_i
\label{ii-1k}
\end{eqnarray}
for each ${\bm\omega}\in\{0,1\}^d$ (including ${\bm\omega}={\bm\omega}_y$
for any $y\in [a,b]$), where the forecast $\f=(f^1,\dots ,f^d)$ can be defined as
\begin{eqnarray}
f^s=\frac{1}{2}-\frac{1}{4}\ln\frac{\sum_{i=1}^N q_i e^{-2(f^s_i)^2}}
{\sum_{i=1}^N q_i e^{-2(1-f^s_i)^2}}
\label{forecast-1f}
\end{eqnarray}
for each $1\le s\le d$.

By letting the grid-size $\Delta\to 0$ (or, equivalently, $d\to\infty$) in (\ref{appr-1}),
(\ref{ii-1k}), where ${\bm\omega}={\bm\omega}_y$, and in (\ref{appr-2}), we obtain
for any $y\in [a,b]$,
\begin{eqnarray}
e^{-\frac{2}{(b-a)}\CRPS(F,y)}\ge\sum_{i=1}^N e^{-\frac{2}{b-a}\CRPS(F_i,y)}q_i,
\label{ii-1kk}
\end{eqnarray}
where $F(u)$ is the limit form of (\ref{forecast-1f}) defined by
\begin{eqnarray}
F(u)=\frac{1}{2}-\frac{1}{4}\ln\frac{\sum_{i=1}^N q_i e^{-2(F_i(u))^2}}
{\sum_{i=1}^N q_i e^{-2(1-F_i(u))^2}}
\label{forecast-2ba}
\end{eqnarray}
for each $u\in [a,b]$.

The inequality (\ref{ii-1kk}) means that the loss function
$\CRPS(F,y)$ is $\frac{2}{b-a}$-mixable.~QED 

Let us specify the protocol 2 of the game with probabilistic predictions for case
when the rule (\ref{forecast-2}) for AA is used.

\smallskip

{\bf Protocol 3}

{\small
\medskip\hrule\hrule\medskip

Define $w_{i,1}=\frac{1}{N}$ for $1\le i\le N$.

\medskip
\noindent \hspace{2mm}{\bf FOR} $t=1,\dots ,T$
\begin{enumerate}
\item Receive the expert predictions -- the probability distribution
functions $F_{i,t}(u)$, where $1\le i\le N$.
\item Present the learner forecast -- the probability distribution
function $F_t(u)$:
\begin{eqnarray}
F_t(u)=\frac{1}{2}-\frac{1}{4}\ln\frac{\sum_{i=1}^N w^*_{i,t}e^{-2(F_{i,t}(u))^2}}
{\sum_{i=1}^N w^*_{i,t}e^{-2(1-F_{i,t}(u))^2}},
\label{forecast-2b}
\end{eqnarray}
where $w^*_{i,t}=\frac{w_{i,t}}{\sum_{j=1}^N w_{j,t}}$.
\item 
Observe the true outcome $y_t$ and compute the score
$\CRPS(F_{i,t},y_t)$ for the experts $1\le i\le N$
and the score $\CRPS(F_t,y_t)$ for the learner.
\item
Update the weights of the experts $1\le i\le N$
\begin{eqnarray}
w_{i,t+1}=w_{i,t}e^{-\frac{2}{b-a}\CRPS(F_{i,t},y_t)}
\label{weight-up-3}
\end{eqnarray}
\end{enumerate}

\noindent \hspace{2mm}{\bf ENDFOR}
\medskip\hrule\hrule\medskip
}
\smallskip

The performance bound of algorithm defined by Protocol 3 is presented
in the following theorem.
\begin{theorem}\label{theorem-2}
For any $i$
\begin{eqnarray}
\sum\limits_{t=1}^T\CRPS(F_t,y_t)\le\sum\limits_{t=1}^T \CRPS(F_{i,t},y_t) +
\frac{b-a}{2}\ln N
\label{h-t-2a}
\end{eqnarray}
for each $T$.
\end{theorem}
{\it Proof.} The bound (\ref{h-t-2a}) is a direct corollary of the regret analysis
of Section~\ref{app-1} and the bound (\ref{prop-1}).~QED 

The square loss function is also $\eta$-exponential concave for $0<\eta\le\frac{1}{2}$
(see~\citet{cesa-bianchi}). In this case (\ref{forecast-2b}) can be replaced with
the forecast WA
\begin{eqnarray}
F_t(u)=\sum\limits_{i=1}^N w^*_{i,t} F_{i,t}(u),
\label{forecast-2a}
\end{eqnarray}
where $w^*_{i,t}=\frac{w_{i,t}}{\sum\limits_{j=1}^N w_{j,t}}$ are normalized weights.
The corresponding weights are computing recursively
\begin{eqnarray}
w_{i,t+1}=w_{i,t}e^{-\frac{1}{2(b-a)}\CRPS(F_{i,t},y_t)}.
\label{exp-concave-1}
\end{eqnarray}
Using results of~\citet{Kaln2017} (presented in Section~\ref{app-1}), we conclude that
in this case the bound (\ref{h-t-2a}) can be replaced with
\begin{eqnarray*}
\sum\limits_{t=1}^T {\rm \CRPS}(F_t,y_t)\le
\sum\limits_{t=1}^T {\rm \CRPS}(F_{i,t},y_t)+2(b-a)\ln N.
\end{eqnarray*}
The proof is similar to the proof of Theorem~\ref{theorem-2}.

\section{Experiments}\label{exp-1}

The proposed rules (\ref{forecast-2b}) for AA and (\ref{weight-up-3}) for WA can be used
in the case when the probability distributions presented by the experts are
given in the closed form (i.e., distributions given by analytical formulas).
For this case, numerical methods can be used to calculate the integrals ($\CRPS$) with
any degree of accuracy given in advance (see also Footnote 9).

In the experiments, we have used Fixed Share modification (see~\citealt{HeW98})
of Protocol 3 and 3a, where we replace the rule (\ref{weight-up-3}) with the two-level scheme
\begin{eqnarray*}
w^\mu_{i,t}=\frac{w_{i,t}e^{-\frac{2}{b-a}\CRPS(F_{i,t},y_t)}}
{\sum\limits_{j=1}^N w_{j,t}e^{-\frac{2}{b-a}\CRPS(F_{j,t},y_t)}},
\\
w_{i,t+1}=\frac{\alpha}{N}+(1-\alpha)w^\mu_{i,t},~~~
\end{eqnarray*}
where $0<\alpha<1$. We do the same for the rule (\ref{exp-concave-1}).
We set $\alpha=0.001$ in our experiments.\footnote{
In this case, using a suitable choice of the parameter $\alpha$, we can obtain
a bound $O((k+1)\ln (TN))$ for the regret of the corresponding algorithm, where
$k$ is the number of switching in the compound experts.}

\subsection{Synthetic data}\label{exper-1}

\begin{figure}[!htb]
\includegraphics[scale=0.32]{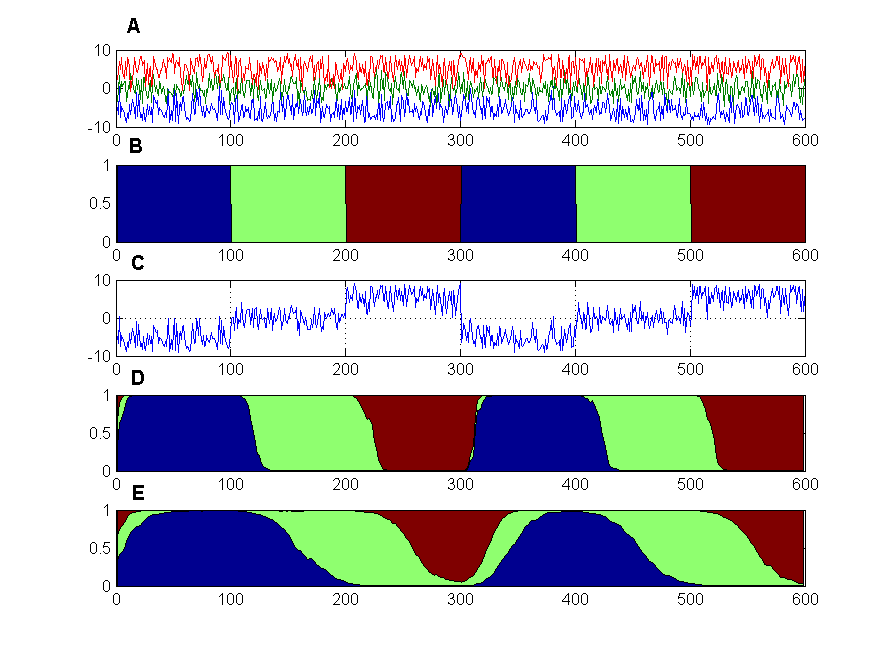}
\includegraphics[scale=0.32]{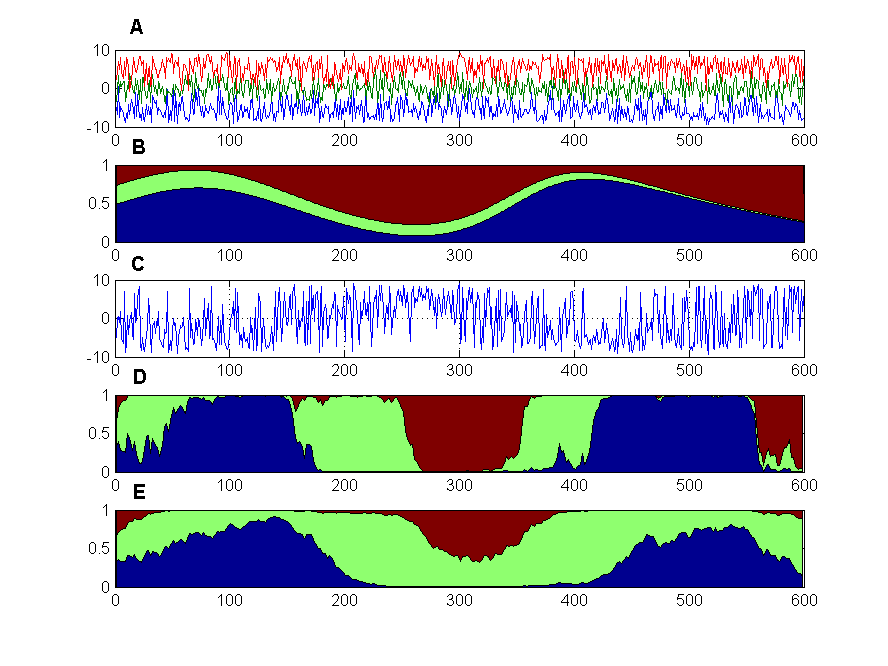}
        \caption{{\small The stages of numerical experiments and the results of
experts' aggregation for two data generation methods (Method 1 -- left, Method 2 - right).
(A) -- realizations of the trajectories for the three data generating distributions;
(B) -- weights of the distributions assigned by the data generating method;
(C) -- sequence sampled from the distributions defined by Method 1 and Method 2;
(D) -- weights of the experts assigned online by the AA using the rule
(\ref{weight-up-3}) and Fixed Share update;
(E) -- weights of the experts assigned online using the rule (\ref{exp-concave-1})
and Fixed Share.
}}\label{fig-2}
          \end{figure}

\begin{figure}[!htb]
\includegraphics[scale=0.32]{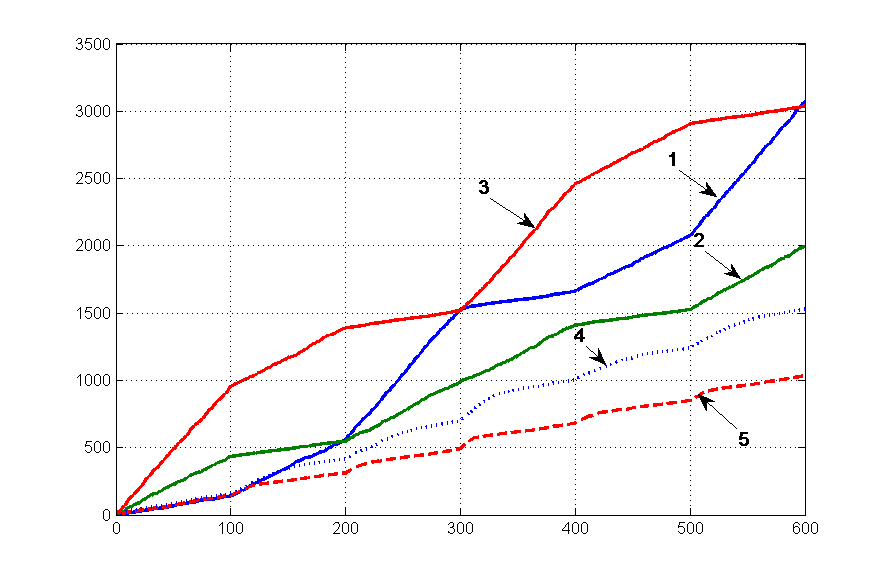}
\includegraphics[scale=0.32]{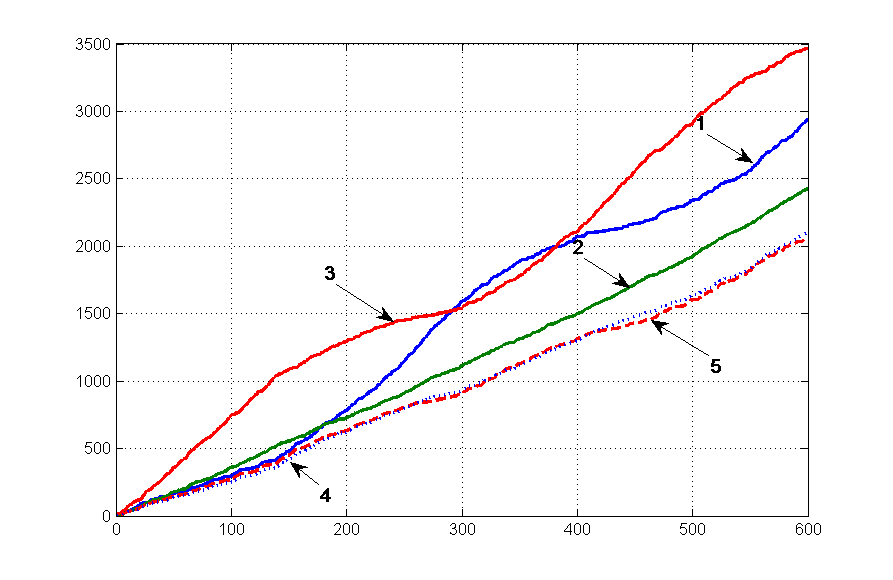}
        \caption{{\small
The cumulated losses of the experts (lines 1-3) and
of the aggregating algorithm for both data generating methods
(Method 1 -- left, Method 2 - right) and for both methods of computing
aggregated forecasts:
line 4 -- for the rule (\ref{forecast-2a}) and line 5 -- for the rule
(\ref{forecast-2b}). We note an advantage of rule (\ref{forecast-2b})
over rule (\ref{forecast-2a}) in the case of data generating Method 1,
in which there is a rapid change in leadership of the data generating distributions.
}}\label{fig-3}
                  \end{figure}

\begin{figure}[!htb]
\includegraphics[scale=0.30]{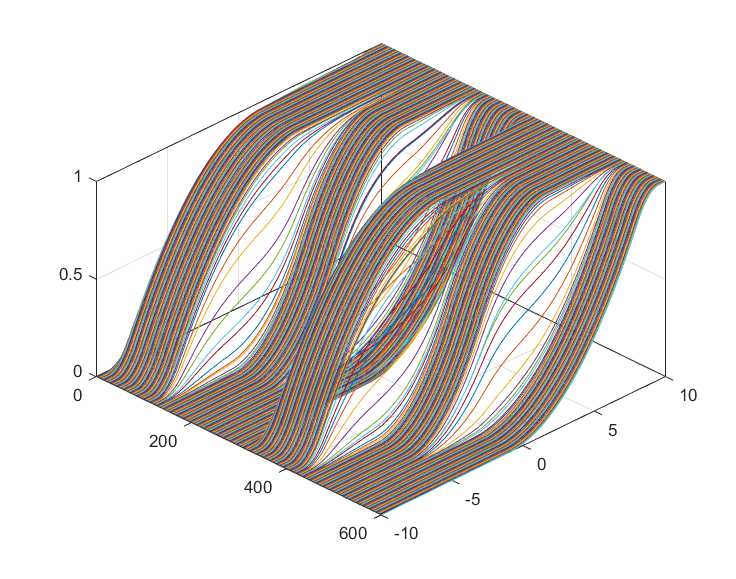}
\includegraphics[scale=0.30]{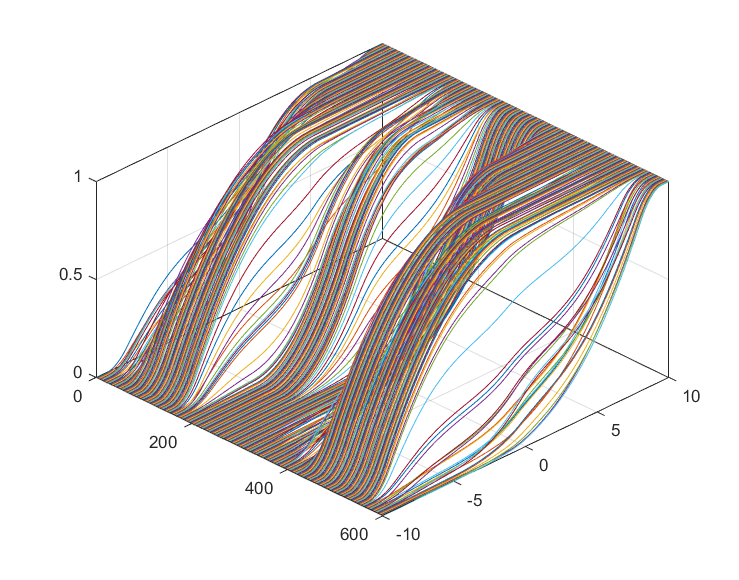}
        \caption{{\small
Empirical distribution functions obtained online as a result of aggregation
of the distributions of three experts by the rule
(\ref{forecast-2b}) for both data generating methods.
}}\label{fig-4}
                  \end{figure}

In this section we present the results of experiments with AA and WA which were performed on
synthetic data. The initial data was obtained by sampling from a mixture of the three distinct
probability distributions with the triangular densities.
The time interval is made up of several segments of the same length, and the weights
of the components of the mixture depend on time. We use two methods of mixing.
By Method 1, only one generating probability distribution is a leader
at each segment (i.e. its weight is equal to one). By Method 2, the weights of the mixture
components vary smoothly over time (as shown in section B of Figure~\ref{fig-2}).

There are three experts $i=1,2,3$, each of which assumes that the time series under
study is obtained as a result of sampling from the probability distribution
with the fixed triangular density with given peak and base.
Each expert evaluates the similarity of the testing point of the series
with its distribution using $\CRPS$ score.

We compare two rules of aggregations of the experts' forecasts: Vovk's AA (\ref{forecast-2b})
and the weighted average WA (\ref{forecast-2a}).

Figure~\ref{fig-2} shows the main stages of data generating
(Method 1 -- left, Method 2 - right) and the results of aggregation of the experts models.  
Section A of the figure shows the realizations of the trajectories of
the three data generating distributions.
The diagram in Section B displays the actual relative weights
that were used for mixing of the probability distributions.
Section C shows the result of sampling from the mixture distribution.
The diagram of Sections D and E show the weights of the experts assigned
by the corresponding Fixed Share algorithm in the online aggregating process
using rules (\ref{forecast-2b}) and (\ref{forecast-2a}).

Figure~\ref{fig-3} shows the cumulated losses of the experts and
the cumulated losses of the aggregating algorithm for both data generating methods
(Method 1 -- left, Method 2 - right) and for both methods of computing the aggregated
forecasts -- by the rule (\ref{forecast-2b}) and by the rule (\ref{forecast-2a}).
We note an advantage of rule (\ref{forecast-2b})
over the rule (\ref{forecast-2a}) in the case of data generating Method 1,
in which there is a rapid change in leadership of the generating experts.

Figure~\ref{fig-4} shows in 3D format the empirical distribution functions obtained
online by Protocol 3 for both data generating methods and the rule (\ref{forecast-2b}).

\subsection{Aggregation of probabilistic predictions with confidence}\label{CRPS-3}

In Section~\ref{exper-2} (below) we present results of numerical experiments
with the real data and when prediction of the experts are supplied by the
levels of confidence. In this case we use a modification of Protocol 3 -- Protocol 3a,
which is presented below.


\smallskip

{\bf Protocol 3a}

{\small
\medskip\hrule\hrule\medskip

Define $w_{i,1}=\frac{1}{N}$ for $1\le i\le N$.

\medskip
\noindent \hspace{2mm}{\bf FOR} $t=1,\dots ,T$
\begin{enumerate}
\item Receive the expert predictions -- the probability distribution
functions $F_{i,t}(u)$ and confidence levels $p_{i,t}$, where $1\le i\le N$.
\item Present the learner forecast -- the probability distribution
function $F_t(u)$ which is defined by the rule (\ref{forecast-2b}) (AA)
or by the rule (\ref{forecast-2a}) (WA), where
$w^*_{i,t}=\frac{p_{i,t} w_{i,t}}{\sum_{j=1}^N p_{j,t} w_{j,t}}$.
\item 
Observe the true outcome $y_t$ and compute the score
$\CRPS(F_{i,t},y_t)$ 
for the experts $1\le i\le N$
and the score
$\CRPS(F_t,y_t)$ 
for the learner.
\item
Update the weights of the (virtual) experts $1\le i\le N$
\begin{eqnarray}
w_{i,t+1}=w_{i,t}e^{-\eta (p_{i,t}\CRPS(F_{i,t},y_t)+(1-p_{i,t})\CRPS(F_t,y_t))},
\label{for-1b-2c}
\end{eqnarray}
where $\eta=\frac{2}{b-a}$ if the rule (\ref{forecast-2b}) is used and
$\eta=\frac{1}{2(b-a)}$ for the rule (\ref{forecast-2a}).

\end{enumerate}

\noindent \hspace{2mm}{\bf ENDFOR}
\medskip\hrule\hrule\medskip
}
\smallskip

The performance of this algorithm is presented by the inequality (\ref{mTT-1afaa})
from Theorem~\ref{main-3fa}, where $h_t=\CRPS(F_t,y_t)$,
$l_{i,t}=\CRPS(F_{i,t},y_t)$ and $\eta=\frac{2}{b-a}$ if the rule (\ref{forecast-2b})
for computing the learner's forecast was used and $\eta=\frac{1}{2(b-a)}$ if the rule
(\ref{forecast-2a}) was used.

\subsection{Probabilistic forecasting of electrical loads}\label{exper-2}

The second group of numerical experiments on probabilistic forecasting
were performed with the data of the 2014 (GEFCOM 2014,Track Load) competition conducted
on the Kaggle platform (\citealt{TH2016}).

The main unit of the training sample includes data on hourly electrical load
for 69 months from January 2005 to September 2010 and data on hourly
temperature measurements during 117 month period.
Databases are available at \url{http://www.kaggle.com/datasets}.
\begin{figure}[!htb]

\includegraphics[scale=0.22]{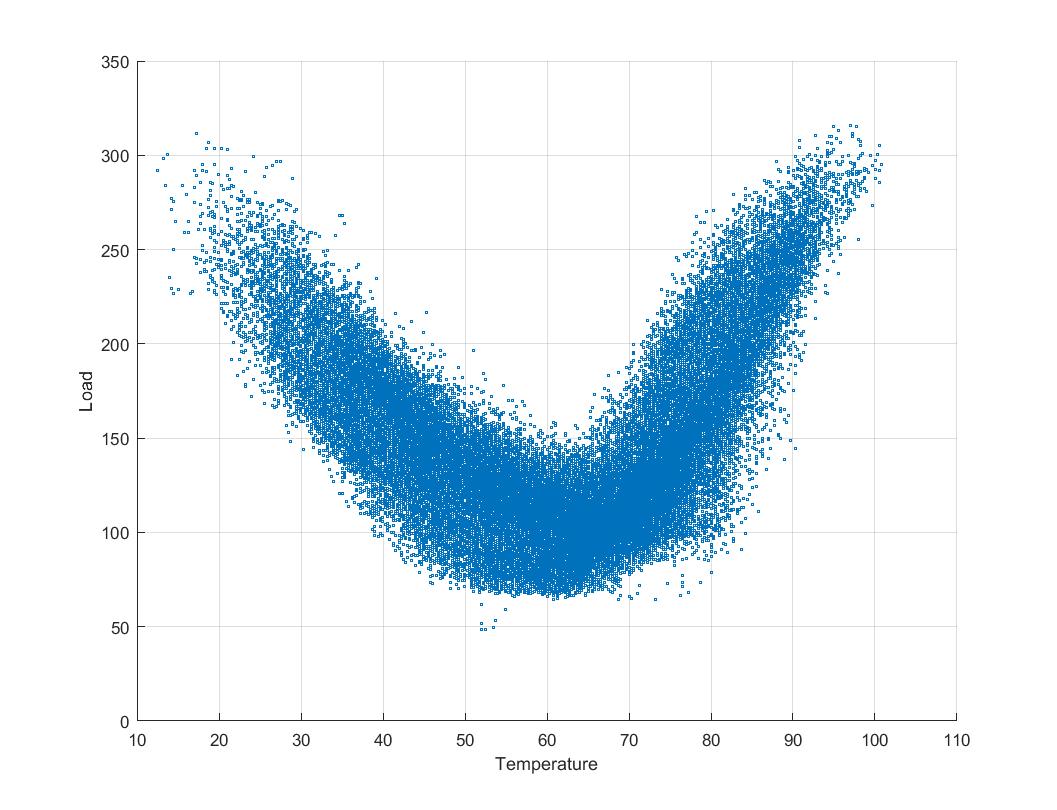}
\includegraphics[scale=0.22]{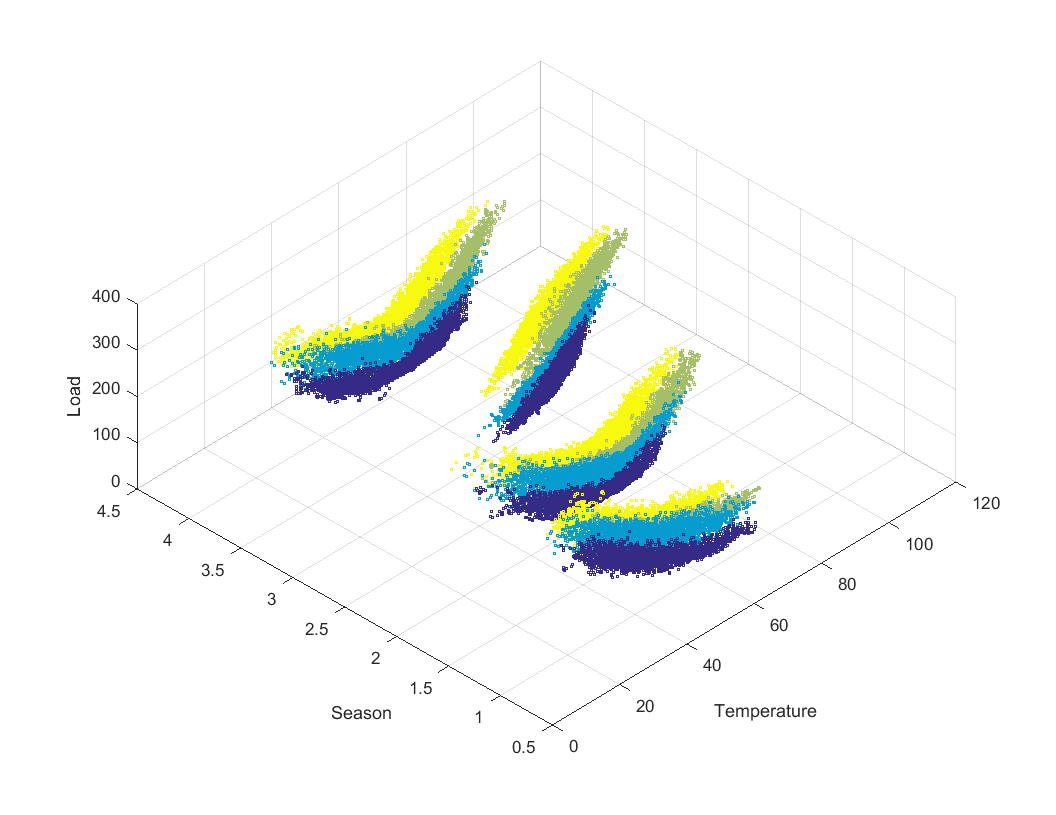}

          \caption{{\small Daily temperature and electrical load paths for all days
from January 2001 to September 2010. Left figure -- all data; right figure -- data
grouped by seasons (Winter, Spring, Summer, Autumn) and time of day marked in color
(Night, Morning, Day, Evening).
}}\label{fig-5e}
\end{figure}

The scatter diagrams ``Load - Temperature'' for several sets of calendar parameters:
(four seasons of the year and four consecutive intervals of day, each for 6 hours)
are presented in Figure~\ref{fig-5e}. The diagrams are constructed according to
the training part of the sample.

Figure~\ref{fig-5e} shows the nature of the relationship between potential predictors
and response.
This data shows the dependence of electrical loads on temperature looking differently
during different seasons and time of day.
For each of the scattering diagrams presented, two temperature intervals can be
distinguished in such a way that within each intervals the point cloud has
a simple ellipsoidal shape. This provides the basis for using a mixture of
normal distributions for the probabilistic forecast of the expected electrical
load according to the short-term temperature forecast.

Scatter patterns on Figure~\ref{fig-5e} can serve as the basis for determining
the pool of the experts. Each of them learns (builds a predictive probabilistic model)
at sample points related to a predefined calendar segment, for example
``Winter$\&$Morning'',  etc. These segments should cover all possible combinations of
calendar indicators present in the data.

A set of 21 specialized experts is defined by dividing the calendar space into areas
where the relationship between temperature and electrical load can be described
by a simple and relatively uniform dependence.
To define an expert a combined sample of historical data consisting of the initial sample
of ``temperatures -- loads'' ensemble, as well as its competence area (season, time of day),
was determined.

The anytime Expert 1 corresponds to the left part of Figure~\ref{fig-5e}, Experts 2-5
correspond to four seasons (see right part of Figure~\ref{fig-5e}). Experts 6-21
correspond to the colored parts of the plots on the right part of Figure~\ref{fig-5e}.
To construct the probability distribution of any expert, we use the method of
Gauss Mixture Models (GMM), which is applied to the corresponding ensemble
of ``temperatures -- loads''.
This probabilistic model is presented as a mixture of two normal distributions.

 \begin{figure}[!htb]
\centering\includegraphics[height=60mm,width=135mm,clip]
        {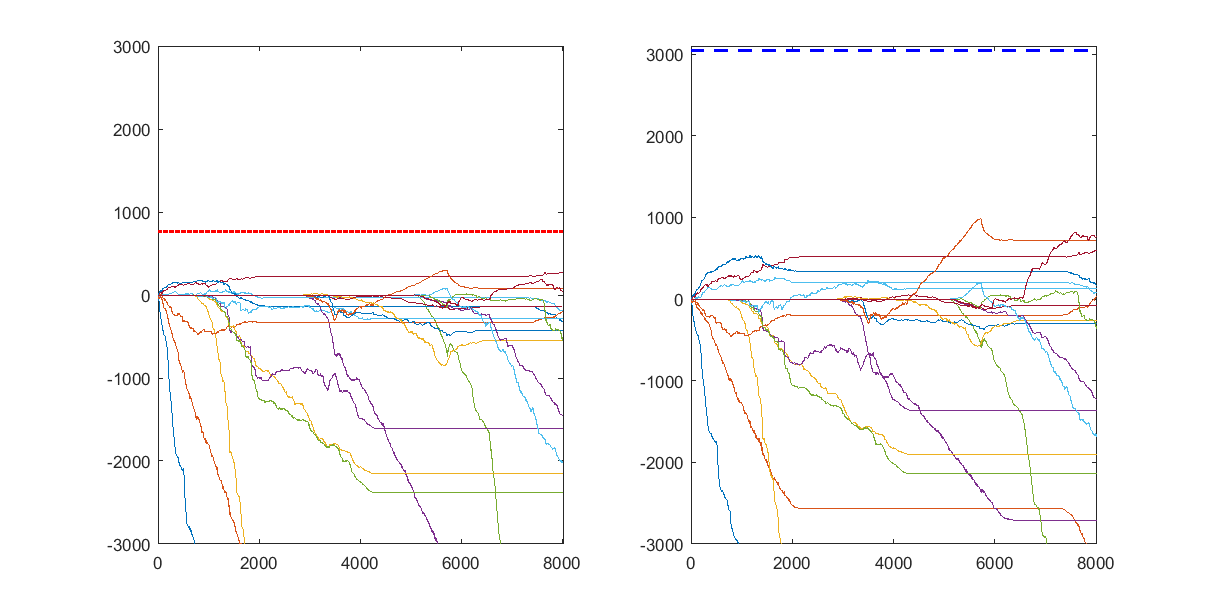}
          \caption{{\small Discounted regret curves for AA (left) and WA
(right) with respect to each of 21 specialized experts. The dotted lines
above represents the theoretical bounds for the regret.
}}\label{fig-6e}
\end{figure}



\begin{figure}[!htb]

\includegraphics[scale=0.28]{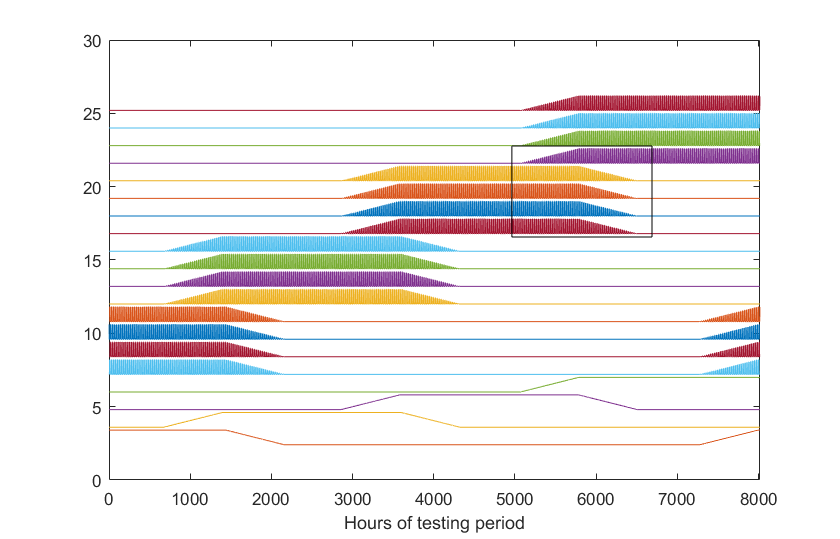}
\includegraphics[scale=0.28]{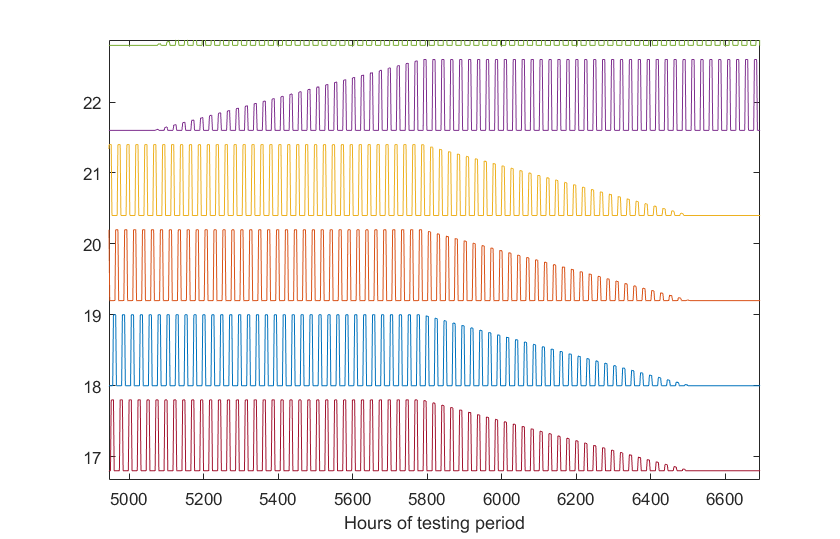}

          \caption{{\small
Left part: confidence levels for for Experts 2-5 (season experts) and 6--21
(``season$\&$time of day''). Right part: enlarged fragment.
}}\label{fig-10e}
\end{figure}

\begin{figure}[!htb]

\includegraphics[scale=0.45]{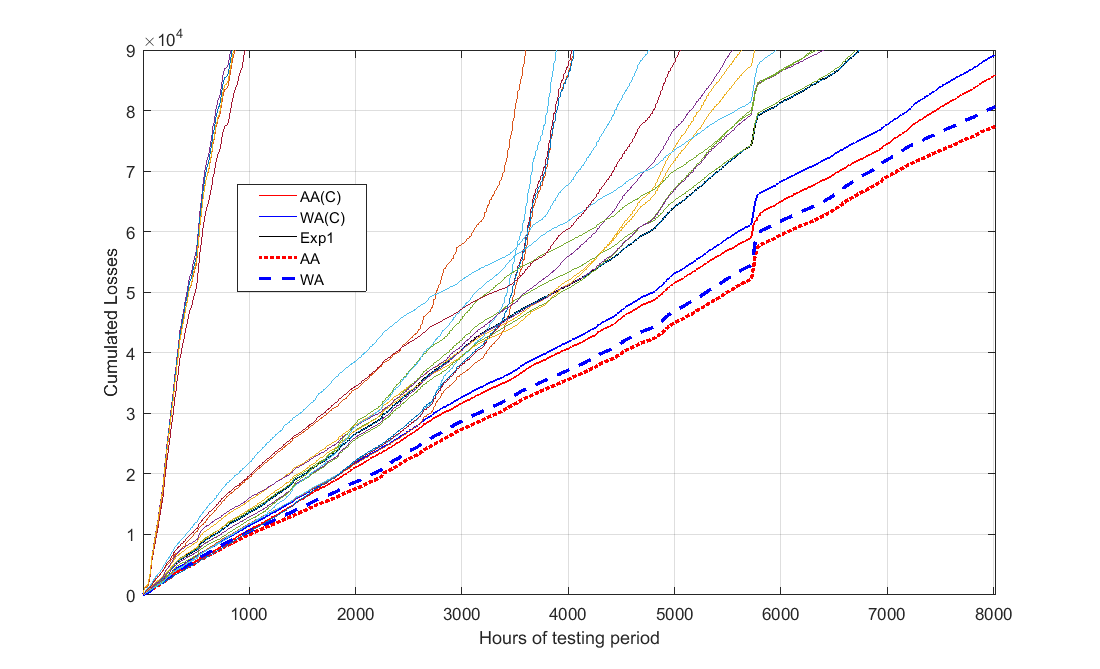}

          \caption{{\small {\bf Comparative study
of learning with/without specialization of the experts.}
1) Cumulated losses of all 21 specialized experts
working any time; 2) results of their aggregation by AA and WA,
where confidence levels of the experts are set to 1;
3) results of aggregation by AA and WA of specialized experts,
where theirs confidence levels are taken into account.
}}\label{fig-7e}
\end{figure}

\begin{figure}[!htb]

\includegraphics[scale=0.37]{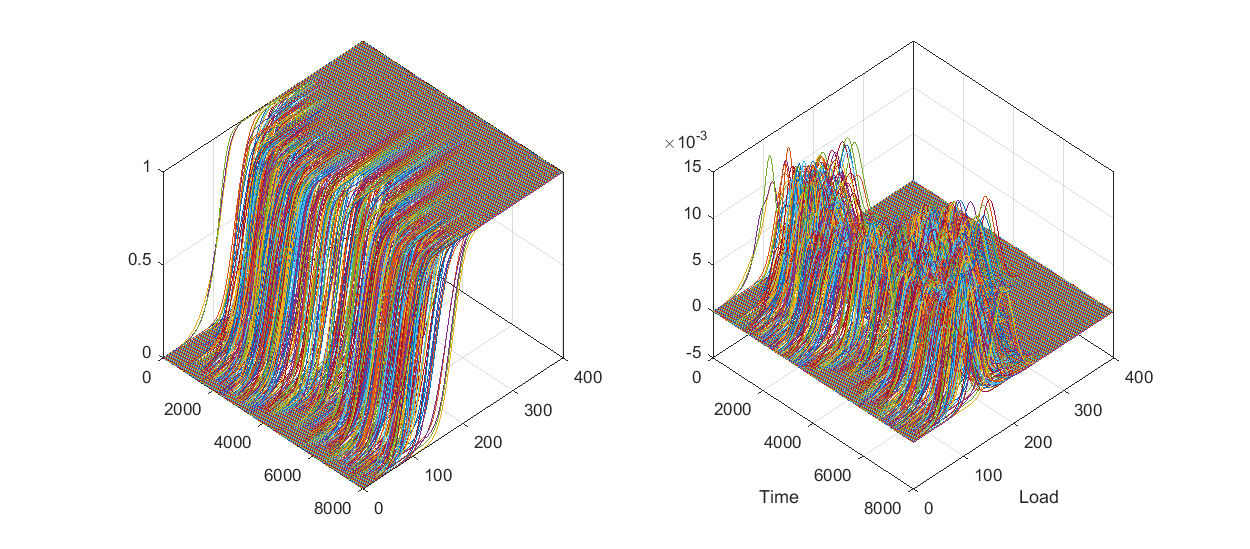}
          \caption{{\small Time changing of probability forecasts -- probability
distribution functions (left) and of the corresponding densities (right).}}\label{fig-11e}
\end{figure}
We consider a particular forecasting problem -- the short-term forecasting of
a probability distribution function for one hour in advance. The scope of each expert
is determined by its confidence function. When forecasting, the expert's smooth areas
of competence are chosen wider than those areas in which this expert was trained.
Thus, each expert competes with other experts working at overlapping intervals
using the corresponding algorithm for combining experts with confidence
levels from Section~\ref{CRPS-3},
like it was done for computing the pointwise forecasts by~\citet{VyT2019-1}.

The discounted regret curves $T\to\sum\limits_{t=1}^T p_{i,t}(h_t-l_{i,t})$
for AA and WA with respect to each of 21 specialized experts are presented in
Figure~\ref{fig-6e}.
The dotted lines above represent the theoretical bounds
for the regret (see the inequality (\ref{mTT-1afaa})).


To justify the role of confidence parameters, the comparative experiments were conducted.
During the first experiment, all confidence values for each expert were equal to 1.
In the second experiment, AA and WA algorithms used the experts predictions within
the levels of their confidence.

The results of both experiments are presented in Figure~\ref{fig-7e}.
The cumulated losses of all 21 specialized experts working any time are presented
in this figure. Cumulative loss curves show that specialized experts, which were trained
only for certain types of data, quickly lose their effectiveness in other types
of data areas and generally suffer large losses. An exception is Expert 1, who trained
on all types of data.

The results of two methods of aggregation of these experts by AA and WA
are also presented in Figure~\ref{fig-7e}.
In the first method of aggregation, confidence levels
of all experts were equal to 1.

In the second method, algorithms AA and WA use specialized experts,
where theirs confidence levels are set externally.
They correspond to the training intervals of specialized experts, but are somewhat
wider and monotonically decrease to zero outside these intervals
(see example in Figure~\ref{fig-10e}).
Confidence levels were not optimized in this experiment.
The results of aggregation by AA and WA of specialized experts are also presented
in Figure~\ref{fig-7e}. 
The results of the experiments show that the use confidence levels
of specialized experts increases the efficiency of the process of online adaptation.
These results also show that AA in all experiments slightly 
outperforms WA.

Time changes of probability forecasts (probability distribution functions) and of
the corresponding densities are presented on Figure~\ref{fig-11e}.

\section{Conclusion}

In this paper the problem of aggregating the probabilistic forecasts is considered.
In this case, a popular example of proper scoring rule for continuous
outcomes is the continuous ranked probability score $\CRPS$. 

We present the theoretical analysis of the continuous ranked probability score
$\CRPS$ in the prediction with expert advice framework and
illustrate these results with computer experiments.

We have proved that the $\CRPS$ loss function is mixable and and then all machinery
of the aggregating algorithm by~\citet{VoV98} can be applied.
The proof is an application of prediction of packs by~\citet{Kaln2017}: the probability
distribution function can be approximated by a piecewise-constant function and further
the method of aggregation of the generalized square loss function have been used.

Basing on mixability of $\CRPS$, we propose two methods for calculating the predictions
using the aggregating algorithm (AA) and the weighted average of
forecasts of the experts (WA).
The time-independent upper bounds for the regret were obtained for both methods.


The proposed methods are closely related to the so called ensemble forecasting
(\citealt{TMB2017}).
In practice, the output of physical process models usually not probabilities,
but rather ensembles. Ensemble forecasts are based on a set of physical models.
Each model may have its own physical formulation, numerical formulation and input data.
An ensemble is a collection of model trajectories,
generated using different initial conditions of model equations.
Consequently, the individual ensemble members represent likely scenarios of the future
physical system development, consistent with the currently available incomplete
information. In this case, the aggregation methods of the corresponding 
ensemble based probability distribution functions may be useful.


We have presented the results of numerical experiments based on the proposed
methods and algorithms. These results show that two methods of computing
forecasts AA and WA lead to similar empirical cumulative losses while the
rule (\ref{forecast-2b}) results in four times less regret bound than (\ref{forecast-2a}).
We note a significantly best performance of method AA (\ref{forecast-2b})
over method WA (\ref{forecast-2a}) in the case where
there is a rapid change in leadership of the experts.
This difference has been demonstrated in numerical experiments.

\section*{Acknowledgement}

The authors are grateful to Vladimir Vovk and Yuri Kalnishkan for useful discussions
that led to improving the presentation of the results.
This paper is an extended version of the conference COPA--2019
(Conformal and Probabilistic Prediction with Applications) paper by~\citet{VyT2019-2}.
This work was partially supported by Russian Science Foundation, project 20-01-00203.

\end{document}